# Diffusion Attention Expert Model for Predicting and Semi-automatic Localizing STAS in Lung Cancer Histopathological Images

Liangrui Pan(潘良睿)[1+*], Jiadi Luo[2,3+], Yuxuan Xiao[4+], Chenchen Nie[5], Xiaoshuai Wu[1], Songqing Fan[2,3+], Ling Chu[6], Manqiu Li[7], Rongfang He[8], Zhenyu Zhao[4], Ruixing Wang[9], Shulin Liu[10], Yiyi Liang[11], Xiang Wang[4], Qingchun Liang[2,3*], Shaoliang Peng[1*]

[1]College of Computer Science and Electronic Engineering, Hunan University, Changsha, 410082, China

[2]Department of Pathology, The Second Xiangya Hospital, Central South University, Changsha, 410011, Hunan, China

[3]Hunan Clinical Medical Research Center for Cancer Pathogenic Genes Testing and Diagnosis, Changsha, Hunan, 410011, China

[4]Department of Thoracic Surgery, The Second Xiangya Hospital, Central South University, Changsha, 410011, Hunan, China

[5]Department of pathology, Hunan Cancer Hospital, The Affiliated Cancer Hospital of Xiangya School of Medicine, Central South University, Changsha, 410013, Hunan, China.

[6]Department of Pathology, The Third Xiangya Hospital, Central South University, Changsha, 410013, Hunan, China

[7]Department of Pathology, First People's Hospital of Pingjiang County, Pingjiang County, 414508, Hunan, China

[8]Department of Pathology, the First Affiliated Hospital, Hengyang Medical School, University of South China, Hengyang, Hunan, China

[9]Department of Radiology, The Second Xiangya Hospital of Central South University, Changsha, Hunan Province, China

[10]Department of Radiology, Xiangya Hospital, Central South University, Changsha, Hunan, China 410008

[11]Oncology Department and State Key Laboratory of Systems Medicine for Cancer of Shanghai Cancer Institute, Renji Hospital, School of Medicine, Shanghai Jiaotong University, Shanghai, 200127, China

[+]: equally contributing

*To whom correspondence should be addressed. panlr@hnu.edu.cn, 503079@csu.edu.cn or slpeng@hnu.edu.cn

## ABSTRACT

Accurate intraoperative and postoperative diagnosis of spread through air spaces (STAS) is essential for guiding surgical decisions and postoperative management in lung cancer. However, histopathological assessment is labor-intensive and is prone to missed or incorrect diagnoses. We propose a Diffusion Attention Expert Model (DAEM) to detect STAS in frozen sections (FSs) and paraffin sections (PSs). Its diffusion attention expert module leverages full attention aggregation to learn multi-scale features from histopathological images, while a dual-branch architecture strengthens multi-scale feature representation. On an internal dataset, DAEM achieves AUCs of 0.8946 for FSs and 0.9112 for PSs. Validation on external multi-center datasets from eight institutions demonstrates strong generalizability and interpretability. Using tumor microenvironment (TME) features in PSs, we further enable semi-automatic measurement of STAS location and its distance from the primary tumor. Several quantitative TME metrics are identified as potential biomarkers for STAS, including micropapillary-type STAS. Overall, DAEM offers a clinically actionable framework for STAS assessment by enabling accurate and interpretable detection on FSs and PSs, supporting postoperative risk stratification through quantitative TME-based analysis.

## INTRODUCTION

The phenomenon of spreading through alveolar spaces (STAS) was first discovered and described in 2013 [1]. In the 2015 World Health Organization (WHO) pathological classification, STAS was officially proposed as a new invasive pattern of lung cancer [2]. This definition clearly states that STAS is characterized by micropapillary clusters, solid tumor nests, or isolated cancer cells that extend beyond the tumor's edge and invade the alveolar spaces of the otherwise normal lung tissue [3]. In simple terms, it refers to the detection of tumor cell (cluster) tracks in the alveolar areas of the lung, distant from the main tumor lesion. STAS is relatively common in lung adenocarcinoma, with multiple studies showing an incidence rate of over 30% [3–7]. STAS is more likely to occur in smokers and in lung adenocarcinomas with high-grade histological types, such as those with a predominantly sieve-like, solid, or micropapillary morphology, especially with a significant association with micropapillary growth patterns [8]. STAS is an independent predictor of poor prognosis in lung cancer [9,10]. Patients with STAS exhibit significantly higher postoperative recurrence rates and lung cancer–specific mortality [9]. For example, patients with stage I lung adenocarcinoma who are STAS-positive have markedly worse recurrence-free survival (RFS) and overall survival (OS) compared with those who are STAS-negative [11]. A study involving 785 patients reported that the 3-year RFS and OS rates in STAS-positive patients decreased to 69.6% and 58.4%, respectively, whereas in STAS-negative patients they were 82.7% and 89% [12]. For early-stage lung cancer patients with STAS, lobectomy provides better prognostic outcomes than sublobar resection [2]. After sublobar resection, the recurrence risk in STAS-positive patients increases significantly, whereas lobectomy allows for more complete removal of tumor cells, thereby reducing local recurrence. Furthermore, Chinese Medical Association guideline for clinical diagnosis and treatment of lung cancer (2025 edition) recommends multidisciplinary evaluation for postoperative adjuvant therapy in stage IB patients. It emphasizes that those with high-risk factors such as STAS, vascular invasion, visceral pleural invasion, or palliative resection may require adjuvant chemotherapy or targeted therapy to further reduce the risk of recurrence. Therefore, the diagnosis of STAS plays a crucial role in guiding intraoperative decisions regarding the surgical approach and extent of resection in early-stage lung cancer, and is closely associated with postoperative recurrence, poor prognosis, and the need for adjuvant therapy [9,13–15], [16].

The diagnosis of STAS requires pathologists to perform histopathological evaluation of intraoperative frozen sections (FSs) and paraffin-embedded tissue sections in lung cancer patients. However, the diagnostic process of histopathological images is labor-intensive, time-consuming, and prone to misdiagnosis and missed diagnoses. This makes the rapid and accurate diagnosis of STAS in intraoperative frozen sections and the large-scale accurate diagnosis in paraffin sections (PS) particularly challenging [17]. The evaluation of STAS in FSs can guide the choice of surgical procedure for thoracic surgeons [18,19]. For instance, studies have shown that in T1-stage lung adenocarcinoma patients with STAS-positive results, the recurrence rate after sublobar resection is significantly higher than after lobectomy [9,20,21]. However, the preparation of FSs often leads to morphological changes such as cell folding and distortion, as well as reduced staining quality, which affects the accuracy of STAS pathological diagnosis [22]. Furthermore, some studies have found that tumor cells can shift position due to mechanical forces from the blade, creating artificial artifacts, which are referred to as "spread by the blade" [23,24]. According to relevant studies, the sampling of FSs often fails to adequately cover the alveolar spaces around the tumor, leading to lower sensitivity and higher specificity for detecting STAS [25–27]. Therefore, diagnosing STAS through FSs frequently results in misdiagnosis or missed diagnoses, requiring further diagnosis using PS to confirm the presence of STAS. Clearly defining the STAS status in postoperative pathology reports allows for risk stratification of patients into high- and low-risk categories, providing a basis for subsequent adjuvant treatment decisions. The NCCN 2025 guidelines have classified STAS as a high-risk factor for adjuvant chemotherapy, recommending consideration of adjuvant chemotherapy for stage I patients with high-risk pathological features (such as STAS, micropapillary/solid components) [28]. Therefore, the accurate diagnosis of STAS in PSs, combined with multidisciplinary collaboration in treatment, can maximize the survival benefits for lung adenocarcinoma patients. It is worth noting that there is a significant gap in diagnostic skills between pathologists in regions with developed medical resources and those in less-developed areas, especially in many hospitals in underdeveloped regions where the provision of STAS diagnostic reports is still lacking [29]. With the continued rise in the incidence and mortality rates of lung cancer in China in recent years, issues such as the shortage of pathologists, uneven distribution of medical resources, difficulties in training senior pathologists, and the low establishment rate of pathology departments in county-level hospitals have become particularly prominent [30–32]. In this context, achieving large-scale, rapid, objective, and accurate diagnosis of STAS in histopathological images and incorporating it into routine pathology reports has become a challenge.

Deep learning (DL) has shown considerable potential in assisting the diagnosis of STAS in histopathological images. Since 2022, several studies have demonstrated that DL can directly diagnose STAS from digitized hematoxylin and eosin (H&E) images. For instance, a novel graph model based on Ollivier-Ricci curvature utilizes information from the primary tumor margin to enhance both the predictive accuracy and interpretability of STAS [33]. Feng et al. developed a DL-based model, STASNet, for STAS detection, which also computes semi-quantitative parameters related to STAS density and distance [34]. STASNet achieved a detection accuracy of 0.93 at the patch level and an AUC of 0.72–0.78 at the whole slide image (WSI) level for determining the STAS status [34]. Pan et al. proposed an image analysis model based on a feature interaction Siamese graph encoder to predict STAS from lung cancer histopathological images, achieving AUCs of 0.8275 and 0.8829 in frozen and paraffin-embedded test sections, respectively [35]. Gong et al. introduced a multi-field channel transformer (MFCT) detection model that utilizes a cross-attention mechanism to fuse features from multiple fields, thereby enhancing the capacity of small fields to capture contextual information and improving model accuracy [36]. MFCT obtained F1 scores of 73.4% on the public Ocelot dataset and 85.2% on a custom STAS dataset [36]. The aforementioned methods have not undergone extensive experimental validation on large-scale external datasets to assess their efficacy and generalizability. This lack of substantial validation is a common issue faced by current models. Although deep learning-based approaches have shown potential in STAS detection, their accuracy has yet to meet the standards required for precise pathological diagnosis. Therefore, enhancing the multicenter diagnostic performance of deep learning-based STAS detection remains a clinical necessity.

The algorithms mentioned in the literature primarily address the tasks of histopathological image segmentation and classification. For segmentation, supervised learning is performed based on instance-level annotations of cells to predict the categories and locations of cells in the images. The main segmentation models have evolved from early approaches such as U-Net [37] and R-FCN [38], through mid-term models like Mask R-CNN [39] and U-Net++ [40], to more recent methods such as HoVer-Net and HD-Yolo [41,42]. In contrast, image classification algorithms predominantly rely on Multiple Instance Learning (MIL) methods to analyze histopathological images. These models generally consist of a feature extractor and an aggregation module. The feature extractor, typically a pretrained model based on Convolutional Neural Networks (CNN) or Transformer architectures, processes each patch in a WSI and transforms these patches into vector representations, which are then aggregated to yield a prediction for the entire WSI. MIL models can be categorized into two types: one that directly obtains the final bag-level prediction based on instance-level predictions [43–48], and another that aggregates instance-level features to achieve bag-level prediction [49–53]. However, current MIL methods treat individual tiles in a univariate manner during aggregation without considering their context with other tiles, even though both local and global features are crucial for accurate pathological diagnosis [54]. Given the high resolution, multi-scale nature, complex textures, and high heterogeneity of histopathological images, MIL methods have recently started to analyze WSIs based on multi-scale features and tumor microenvironment (TME) characteristics [55].

In this work, we develop the Diffusion Attention Expert Model (DAEM) to diagnose FSs and PSs, and to semi-

automatically measure the distance from the STAS dissemination foci to the primary tumor by analyzing the spatial positions of various cells within the microenvironment. We employ parallel diffusion attention-based expert models to learn the multi-scale pathological features of tumor cells and the TME in WSIs. A supervised contrastive learning strategy is used to optimize the diffusion attention expert modules, enabling DAEM to achieve mean AUC values of 0.894 and 0.9112 for STAS diagnosis in FSs and PSs, respectively, on the internal dataset. Our proposed method significantly outperforms numerous state-of-the-art (SOTA) models. Additionally, we collect and annotate eight external STAS validation cohorts to further evaluate DAEM. Extensive experiments show that DAEM exhibits excellent generalizability and interpretability. Moreover, by leveraging TME features from WSIs, we semi-automatically measure both the position of STAS and its distance from the primary tumor to assist clinical surgery. Analysis of quantitative TME indicators from WSIs with and without STAS reveals that the tumor stroma ratio (STR), immune cell/tumor cell ratio (ITR), and microvessel density (MVD) are potential biomarkers for STAS, and that MVD and the stroma/vessel ratio (SVR) are potential biomarkers for micropapillary STAS. In summary, DAEM not only supports the diagnosis of STAS in WSIs but also enables its semi-automated localization, thereby providing a robust foundation for pathological assessment by pathologists and thoracic surgeons.

## Results

### STAS prediction based on diffusion attention expert model outperforms state-of-the-art methods

We trained the DAEM on a dataset of histopathological images from lung cancer patients at the Second Xiangya Hospital of Central South University (SXH-CSU) to predict and localize STAS (Figure 1). Given that the overall ratio of FSs to PSs in the dataset is approximately 1:5, we partitioned the dataset into five folds using five-fold cross-validation, ensuring that each fold maintained an FSs-to-PSs ratio of about 1:5 (Supplementary Table 1). In each training phase, DAEM was trained on four folds, with one fold used for validation. After completing five-fold cross-validation, we obtained five optimal DAEM models capable of simultaneously diagnosing FSs and PSs, and localizing STAS by analyzing the spatial distribution of cells in the TME and generating heatmaps of the WSIs. First, for FS diagnosis, we found that the model from the fourth fold of the five-fold cross-validation achieved an AUC of approximately 0.9066 in the intra-domain testing (Figure 2a) and a PRC of 0.8682 (Figure 2b). The average intra-domain test AUC of the five-fold cross-validation model was 0.8954 (Figure 2c), while the average intra-domain test PRC was 0.8541 (Figure 2b). Secondly, the model from the fourth fold cross-validation exhibited the best predictive performance, with accuracy, precision, recall, F1 score, specificity, and AUC values of approximately 0.9043, 0.8333, 0.8974, 0.8642, and 0.9066, respectively (Figure 2c). The confusion matrix from the five-fold cross-validation indicates that DAEM has some false negatives, but overall demonstrates good intraoperative diagnostic performance for STAS (Figure 2d). Through analysis of numerous baseline model experiments, DAEM outperformed the CLAM_SB [56], CLAM_MB [56], ABMIL [57]., DSMIL [45], DTFD [52]., IBMIL [58]., ACMIL [59], PatchGCN [60], DGRMIL [61], TransMIL [49], MIHM [62], ILRA [63], and VERN[35] models in terms of accuracy, precision, recall, F1-score, specificity, and AUC (Figure 3, Supplementary Table 1, P value < 0.01; Supplementary Figure 1,2,3,4). Therefore, we conclude that DAEM can accurately diagnose the presence of STAS in FS during surgery.

Next, PS diagnosis is typically a labor-intensive task [64,65]. By analyzing the results of the five-fold cross-validation experiment, we found that the model from the fourth fold achieved an AUC of approximately 0.9328 in the intra-domain testing (Figure 2e) and a PRC of 0.9172 (Figure 2f). The average intra-domain test AUC of the five-fold cross-validation model was 0.9112 (Figure 2g), while the average intra-domain test PRC was 0.8978 (Figure 2f). Secondly, the model from the fourth fold exhibited the best predictive performance, with accuracy, precision, recall, F1 score, specificity, and AUC values of approximately 0.8726, 0.8805, 0.8615, 0.8709, and 0.9112, respectively (Figure 2g). The confusion matrix from the five-fold cross-validation indicates that DAEM demonstrates good diagnostic performance when analyzing STAS in PS (Figure 2h). Through extensive experimental analysis, DAEM outperformed all other SOTA methods in terms of accuracy, precision, recall, F1-score, specificity, and AUC (Figure 3, Supplementary Table 2, P value < 0.05; Supplementary Figure 1,2,3,4). Therefore, we conclude that DAEM can accurately predict the presence of STAS in FS. However, the model from the first fold of the five-fold cross-validation showed lower performance compared to the other four folds, which may be due to a higher proportion of challenging samples or an abnormal proportion of non-STAS samples in the first fold's data. In conclusion, DAEM can accurately predict both FS and PS with clinical-level performance. This is because the diffusion attention mechanism in DAEM effectively extracts multi-scale features from tissue pathology images, and the parallel expert models ensure consistent feature representation. It is worth noting that the inconsistent sample proportions between FS and PS do not affect the model's accuracy in diagnosing both FS and PS simultaneously.

### DAEM demonstrates robust generalization performance in predicting STAS on external independent validation datasets

To study the generalization of the model, DAEM was externally validated on the STAS datasets from the Xiangya Hospital of Central South University (XH-CSU), the Third Xiangya Hospital of Central South University (TXH-CSU), the Affiliated Tumor Hospital of Zhengzhou University (TH-ZZU), Changsha Economic Development Zone Hospital (CJH), the First People's Hospital of Pingjiang County (PCPH), and the First Affiliated Hospital of Nanhua University (FAH-NHU), as well as the STAS datasets from CPTAC_LUAD and TCGA_LUAD. The external validation sets were curated by three senior pathologists, retaining 1-4 WSIs per patient for testing. The image and image feature preprocessing for the multi-center WSIs followed a process similar to that of the XH-CSU WSIs. We used evaluation metrics such as AUC curve, PRC curve, accuracy, precision, recall, F1 score, specificity, and AUC to assess the generalization performance of DAEM. First, DAEM achieved AUC values of 0.8349, 0.8347, 0.856, 0.8642, 0.8417, and 0.8176, and PRC values of 0.8171, 0.8167, 0.7655, 0.763, 0.8159, and 0.8082 for diagnosing STAS in XH-CSU, TXH-CSU, TH-ZZU, FAH-NHU, CJH, and PCPH datasets, respectively (Figure 4a, 4b, 4c, 4d, 4e, 4f). The diagnostic AUC values for the STAS datasets from the high-level medical centers were all above 0.8, indicating that DAEM's overall diagnostic performance exceeded the level of primary pathologists [66]. However, DAEM showed relatively lower PRC values when tested on the multi-center STAS datasets. This was attributed to an imbalance in the positive and negative sample ratios in some of the STAS datasets provided by the multi-centers, or a higher number of false positives, which lowered the precision. Specifically, in XH-CSU, the DAEM model achieved an accuracy of 0.8194, precision of 0.8571, recall of 0.8217, and an F1 score of 0.8342; in TXH-CSU, accuracy was 0.8651, precision was 0.8444, recall was 0.7692, and F1 score was 0.8541; in TH-ZZU, accuracy was 0.8723, precision was 0.8802, recall was 0.8005, and F1 score was 0.8341; in FAH-NHU, accuracy was 0.8555, precision was 0.8017, recall was 0.8625, and F1 score was 0.8564; in CJH, accuracy was 0.8539, precision was 0.7419, recall was 0.8214, and F1 score was 0.7797; in PCPH, accuracy was 0.7922, precision was 0.8333, recall was 0.8255, and F1 score was 0.8140 (Supplementary Table 3). The results of the external validation indicated that DAEM's performance varied across different hospitals. For instance, in XH-CSU, the model had higher precision but slightly lower recall, whereas in FAH-NHU, the recall was higher, but precision was slightly lower. These variations may be attributed to factors such as the quality of equipment, sample characteristics, and the pathologists' experience at each hospital. The diagnostic performance of DAEM on the PCPH dataset was the lowest, possibly due to the lower data quality caused by differences in sampling, staining, and preparation processes at this center, which led to a decrease in DAEM's diagnostic accuracy.

The data from the TCGA and CPTAC projects come from a wide range of sources and are widely used in artificial intelligence research [67–69]. However, they do not contain annotations for STAS. After cross-validation by pathologists and incorporating manually curated STAS annotations, these datasets were used to evaluate the model's generalizability. DAEM achieved AUC values of 0.7957 and 0.8116, and PRC values of 0.7743 and 0.8109 on the TCGA_LUAD and CPTAC_LUAD STAS datasets, respectively (Figures 4g, 4h,4i). These experimental results indicate that DAEM's generalizability on the TCGA_LUAD and CPTAC_LUAD STAS test sets is relatively lower, possibly due to significant differences in WSI data quality compared to the SXH-CSU dataset and the presence of sample imbalances. Although DAEM achieved high accuracy on these datasets, its low recall suggests that many positive cases may have been missed. Therefore, we may need to optimize the data, and employ domain adaptation techniques to further improve DAEM's generalizability on the TCGA_LUAD and CPTAC_LUAD datasets.

Next, the calibration curves of the DAEM model across eight external validation sets demonstrate the model's consistency and variability across different datasets (Supplementary Figure 5). The Brier scores obtained for the DAEM model on the XH-CSU, TXH-CSU, TH-ZZU, FAH-NHU, CJH, PCPH, TCGA_LUAD, and CPTAC_LUAD datasets were 0.142, 0.139, 0.055, 0.081, 0.138, 0.209, 0.128, and 0.195, respectively. From these subplots, it can be observed that while some calibration curves are close to the perfect calibration line, indicating good performance on those validation sets (e.g., TH-ZZU and TCGA_LUAD), there are also significant deviations, especially in certain validation sets (e.g., XH-CSU and PCPH), where the curves show irregular fluctuations or sudden changes, deviating far from the ideal diagonal. This suggests that the model may be overconfident or lacking confidence in its probability predictions on these datasets, particularly in the middle probability range. Some subplots show smoother upward trends in the calibration curves, indicating that the model adapts well to the data in these validation sets. However, there are also subplots with large deviations or unstable fluctuations, possibly reflecting data heterogeneity or overfitting in specific datasets. This inconsistency suggests that while the DAEM model performs well in certain scenarios, its robustness and generalization ability still need further improvement, especially when facing different external validation sets. In summary, although the DAEM model demonstrates good calibration ability on some validation sets, its performance varies significantly across multiple external datasets, emphasizing the need for further optimization and adjustments in diverse data scenarios.

## Semi-automatic Measurement of the Distance from STAS to the Primary Tumor Assisted by the Tumor Microenvironment

Numerous studies have shown that in STAS-positive patients, the distance from the tumor edge to the farthest STAS in histopathological images is closely associated with prognosis, recurrence risk, and treatment outcomes [70,71] [72] [73,74] [75]. To determine the positional relationship of STAS relative to the primary tumor mass, we leveraged the distribution of cancer cells within the TME to precisely evaluate the distance between STAS and the main tumor. First, we applied a DAEM model to identify WSIs containing STAS and excluded those WSIs without STAS. This procedure significantly conserves clinicians' time and enhances the efficiency of pathological diagnosis in clinical practice. Second, we used the HD-Yolo to obtain the spatial locations and labels of all cells in the TME of STAS-containing WSIs [42]. Based on the regions where tumor cells aggregate, we identified the approximate location of the main tumor body and the positions of STAS outside the main tumor. Third, we developed an interactive web-based platform wherein the left panel displays the original WSIs and the right panel illustrates the spatial distribution of cells within the TME. As shown in Figure 5, our platform includes a distance measurement feature that enables clinicians to accurately acquire the Euclidean distance from a point (representing a cancer cell) to a line delineating the primary tumor's boundary. The left panel presents the histopathological imagery of STAS-containing tissues, whereas the right panel shows the cell distribution map of the corresponding WSI. Both views support distance measurement; however, the TME cell map only reflects cell type without morphological detail, potentially leading to imprecise distance estimates. Therefore, users are encouraged to measure the distance using the STAS from both the TME and the WSI, with a recommendation to rely on the WSI-derived measurements for superior accuracy. Pathologists who tested our tool reported that it reliably localizes STAS. Nonetheless, certain limitations remain. First, the segmentation accuracy for different cell types in the TME is not yet optimal, which may cause non-tumor cells to be erroneously identified as tumor cells; thus, pathologists must manually verify cells situated outside the main tumor. Given that the HD-Yolo achieves an approximate accuracy of 80% in cancer cell recognition, a substantial proportion of cells can still be confidently excluded. Second, the irregular borders of the primary tumor complicate the precise quantification of the distance from STAS to the main tumor. In conclusion, although our method effectively estimates the approximate distance between STAS and the primary tumor mass, achieving exact measurements remains a challenging endeavor.

### DAEM's predictions for STAS diagnosis are interpretable

To investigate the interpretability of the trained DAEM model in diagnosing FSs and PSs in SXH-CSU, we utilized visualization of multi-scale heatmaps of WSIs at 20× and 10× magnifications to assist researchers and clinicians in understanding the model's decision-making process. The heatmaps, which were generated based on the importance of each patch (the probability of tumor vs. non-tumor) and the spatial coordinates of the patches, display regions of high and low attention by DAEM on the WSI (Figure 6a-d). First, pathologists manually annotated regions of interest for Non-STAS and STAS cases in both FSs and PSs. Then, using tumor, stroma, and immune cell segmentation from HD-Yolo, we constructed tumor density maps to reflect the spatial distribution of these components in the WSI. Subsequently, we visualized FSs and PSs at different magnifications using DAEM (Figure 6a, 6c). During the DAEM diagnostic process, the model's high-attention areas were primarily concentrated in tumor cell-rich regions, including the main tumor mass and floating tumor cell clusters in the airways, where the heatmaps showed high-confidence red areas. In cases with extensive STAS, the model's heatmap revealed high-attention signals radiating outward from the tumor's edge into the alveolar spaces, directly illustrating the morphological features of tumor cell dissemination along the airways (Figure 6). Multi-scale visualization showed that 20× magnification (which focuses more on cellular atypia, such as nuclear pleomorphism and chromatin color, along with local histological clues) provides more accurate visualizations compared to 10×. However, the high-attention regions for both magnifications were consistent in the same case. The top, middle, and low attention patches (Top 4, Middle 4, and Low 4) corresponded to these key morphological features (Figure 6b, 6d). Notably, due to the limitations of FS sectioning, slides often show compression and folding, leading to overlapping cells and tissues, which causes darker staining and the appearance of tumor-like cellular morphology due to cell overlap (Figure 6). Furthermore, large areas of hemorrhage are more common in FS than PS, resembling the histological morphology of tumor necrosis with red-stained tissue. These characteristics of FS result in slightly poorer visualization compared to PS.

In the group of images diagnosed by pathologists as Non-STAS, the DAEM model's high-attention regions in the 20× magnification WSI were primarily concentrated in the main tumor body and its remaining blood vessels and bronchi (Figure 6a, 6c). In contrast, the high-attention regions in the 10× magnification WSI extended beyond the main tumor body to include a small portion of blood vessels, lymphocytes, and phagocytic tissue cells, due to the similarity in morphology between bronchial mucosal cells, aggregated lymphocytes, and tissue cells with tumor cells, which posed a diagnostic challenge for the model (Figure 6a, 6c). To further analyze DAEM's attention to patches at different levels, we selected the Top 4, Middle 4, and Low 4 attention patches (Figure 6b, 6d). High-attention patches were mainly composed of tumor cells, medium-attention patches were largely proliferating stromal cells and aggregated inflammatory cells, including some hemorrhagic regions, and low-attention patches consisted primarily of normal alveolar tissue and alveolar secretions. Overall, the tumor regions that DAEM focused on were highly consistent with the regions identified by pathologists, demonstrating the model's strong interpretability. In the cases diagnosed as STAS, DAEM’s attention areas also largely matched those of the pathologists. In cases with extensive STAS, the model effectively displayed tumor cell dissemination along the airways. As a result, pathologists can directly search for high-attention regions outside the main tumor to identify STAS. Although DAEM's high-attention regions in the 10× WSI mostly focused on the main tumor body and did not prioritize STAS regions, the visualization results at 20× magnification showed that DAEM learned the pathological features of STAS during the training process, utilizing multi-scale pathological features. In summary, these results suggest that DAEM effectively learns multi-scale pathological features, including those characteristic of STAS, providing pathologists with a model interpretability framework for the diagnostic process.

To analyze the process of the DAEM model diagnosing the presence of STAS in WSIs across multicenter datasets, we visualized multi-scale heatmaps of the diagnostic process using DAEM. First, we had pathologists annotate their regions of interest (Figure 7). Then, for the diagnosis of non-STAS WSIs, the model's heatmaps at 20× and 10× magnifications show that the model primarily focuses on the main tumor mass, which aligns well with the pathologists' annotated regions of interest. However, in terms of finer details, the model's performance across different centers demonstrates both common traits and certain differences. Overall, in multicenter testing, whether at 20× or 10× magnification, DAEM effectively identifies the STAS regions, displaying stable high signals in the heatmaps, indicating the model's consistent ability to recognize tumor cell clusters. Additionally, the model made some misclassifications across the multicenter data, particularly involving muscular large blood vessels, carbon particle deposits, hemorrhage, and bronchial mucosa. From a pathological morphology perspective, muscular large blood vessels resemble fibrosis found in some tumors, hemorrhage is similar to tumor necrosis, and carbon particle deposits and bronchial mucosa resemble nuclei with deep staining in atypical tumor cells. These features, especially at 10× magnification, can be misleading and posed challenges for the model's diagnosis. Furthermore, WSIs from different centers exhibited certain differences. For example, in the XH-CSU, TH-ZZU, and PCPH centers, the staining styles used in slide preparation led to more distinct color differences between tumor regions and normal lung tissue, resulting in better model visualization and higher accuracy in recognizing STAS. In conclusion, despite some misclassifications and differences between multicenter datasets, the DAEM model demonstrates strong diagnostic capability for STAS. The heatmaps showcase the model's precision in identifying tumor boundaries at 20× magnification and its overall control of the main tumor mass at 10× magnification. The model accurately identifies STAS tumor cell clusters at both magnifications. By combining multi-scale visualization results, the strengths of each magnification complement one another, significantly improving the model's interpretability and providing pathologists with more persuasive auxiliary information.

## Pathological characteristics and related mechanisms of tumor microenvironment-assisted analysis of STAS

To explore relevant biomarkers in the STAS microenvironment, we visualized both the heatmaps of WSIs cross-validated and annotated by three pathologists, as well as all cell spatial maps, to quantify TME information. Initially, pathologists manually annotated WSIs based on regions of interest. Subsequently, WSIs were visualized using attention-based heatmaps, which highlighted model focus on the primary tumor area, including peripheral cancer cells and adjacent vasculature. Pathologist validation confirmed that the model also attended to STAS-involved regions (Figure. 8a). In addition, we spatially visualized the distribution and interactions among various cell types within the TME, including tumor cells, stromal cells, immune cells, erythrocytes, macrophages, dead cells, and other cell populations (Figure. 8a). By analyzing the spatial distribution patterns and clustering behavior of these cell types, we derived quantitative metrics reflective of the TME. Given the statistical requirements of t-test analysis, we performed group-wise comparisons of TME metrics between WSIs with and without STAS across datasets with sufficient sample size, including TCGA_LUAD, CPTAC_LUAD, SXH-CSU, and TXH-CSU. In the intergroup analysis of these datasets, we found that STR, ITR, and MVD all showed statistical significance (Figure 8b, c, d). Therefore, we speculate that STR, ITR, and MVD may be closely associated with the occurrence of STAS.

Furthermore, extensive clinical evidence has demonstrated that lung cancer patients with a micropapillary pattern exhibit a higher degree of malignancy compared to those without this histological subtype [8,76,77]. Inspired by these findings, we further examined differences in the TME between STAS with micropapillary subtype and STAS without micropapillary features. Heatmap analysis revealed that DAEM similarly focused on primary tumor regions, as well as STAS-involved and necrotic areas in these WSIs (Figure. 8e). To analyze the reasons behind the risk differences between micropapillary-type STAS and non-micropapillary-type STAS, we conducted Kruskal-Wallis tests on quantitative indicators within the TME. Among the TME quantitative indicators in WSIs from non-STAS, non-micropapillary STAS, and micropapillary STAS groups, we found that MVD and SVR showed statistically significant differences in the TXH-CSU ($2.9e^{-2}$, positive correlation; $1.6e^{-3}$, positive correlation), SXH-CSU ($2.25e^{-8}$, positive correlation; $6.34e^{-12}$, positive correlation), CPTAC_LUAD ($5.3e^{-4}$, positive correlation; $2.3e^{-4}$, positive correlation), and TCGA_LUAD ($6.6e^{-3}$, positive correlation; $4.5e^{-2}$, positive correlation) datasets. Therefore, we speculate that MVD and SVR may serve as biomarkers for the micropapillary subtype of STAS. In summary, the pathological subtype of STAS is closely related to its tumor microenvironment.

To evaluate the impact of biomarkers on the prognosis of STAS-positive patients, we aggregated and averaged the TME quantitative indicators across all WSIs from STAS and Non-STAS lung cancer patients. Patients were then stratified into high-expression and low-expression groups for STR, ITR, and MVD based on the median values, and Kaplan–Meier survival curves were plotted. Survival differences between groups were assessed using the log-rank test. As shown in Figure. 8h, the Kaplan–Meier survival curves for patients in the SXH-CSU, CPTAC_LUAD, and TCGA_LUAD cohorts demonstrated that patients in the STR high-expression group had significantly worse overall survival than those in the low-expression group, with log-rank $P$ values of 0.0032, 0.04, and 0.049, respectively. Similarly, patients in the ITR high-expression group had significantly poorer overall survival compared to the low-expression group, with log-rank $P$ values of 0.0109, 0.043, and 0.043. For MVD, the high-expression group also showed significantly reduced overall survival, with log-rank $P$ values of 0.0096, 0.019, and 0.034. These results suggest that STR, ITR, and MVD may serve as potential prognostic biomarkers for STAS-positive lung cancer patients. Next, to assess the effect of these biomarkers on recurrence, we utilized recurrence data from patients in the CPTAC_LUAD and TCGA_LUAD cohorts and plotted Kaplan–Meier recurrence-risk curves using the same stratification method (Figure. 8i). We found that STR, ITR, and MVD effectively distinguished STAS-positive lung cancer patients into high- and low-risk recurrence groups. Therefore, STR, ITR, and MVD may represent potential recurrence-risk biomarkers. Furthermore, to evaluate the impact of biomarkers in micropapillary-type STAS patients, we generated Kaplan–Meier survival curves based on survival and recurrence data from STAS-positive patients in the CPTAC_LUAD and TCGA_LUAD datasets. Both survival and recurrence-risk curves clearly stratified micropapillary STAS patients into distinct risk groups, with all log-rank P values < 0.05, indicating statistical significance (Supplementary Table 3, Supplementary Figure 6). These findings suggest that ITR and MVD may serve as potential prognostic and recurrence biomarkers for micropapillary-type STAS patients.

## The workflow based on DAEM can be involved in clinical-grade auxiliary diagnosis.

Histopathological imaging is the gold standard for diagnosing STAS. Therefore, diagnosing STAS requires experienced pathologists to analyze and identify it accurately. Some studies have reported that manual diagnosis of STAS has relatively low accuracy, with an AUC not exceeding 0.8 [27,66]. Notably, the DAEM model has been extensively validated across eight external STAS datasets worldwide, demonstrating the reliability of its performance. Its diagnostic AUC for FS and PS reached 0.8946 and 0.9112, respectively. This indicates that DAEM can be used for preliminary screening of FS and PS to assist junior pathologists in accurately determining the presence of STAS in WSIs. In particular, in regions and countries with imbalanced medical resources, our investigation revealed that accurate STAS diagnoses are still absent from lung cancer diagnostic reports. However, numerous studies have shown that postoperative recurrence in STAS-positive patients is closely related to the surgical resection range. Local resection of lung cancer with STAS characteristics may increase the recurrence rate [78,79]. Therefore, intraoperative STAS diagnosis to guide surgical decisions is urgently needed.

We propose a workflow for the diagnosis and localization of STAS (Figure 9). First, all FS and PS are digitized using a scanner, a process that takes approximately 2 minutes (depending on the scanner). Since the detection of STAS in FSs is easily affected by tissue sampling, intraoperative FS examination is primarily aimed at determining whether STAS is present. This process requires a rapid and accurate diagnosis of STAS within WSIs, with results communicated promptly to pathologists and thoracic surgeons. In our practical workflow, the time required for AI-assisted diagnosis was shorter than that of multiple pathologists making the same determination. Next, for postoperative PSs, pathologists can utilize the DAEM model to perform large-scale STAS diagnosis and further determine the distance between STAS and the primary tumor within WSIs. Given the strong correlations of STAS with EGFR and ALK mutations [80,81]. STAS diagnosis may influence decisions regarding adjuvant therapy. Both pathologists and oncologists can therefore leverage this workflow to accelerate STAS diagnosis and localization, and to formulate appropriate postoperative treatment strategies. To this end, we provide a free platform for diagnosing STAS and measuring the distance from STAS to the primary tumor, which serves as an auxiliary tool for pathologists. Importantly, for data security and privacy protection, we also recommend downloading the DAEM model for local testing, which allows our backend to avoid collecting user data.

## Discussion

This study holds significant clinical implications. Firstly, we developed a DAEM to diagnose STAS in both FSs and PSs, utilizing the TME for semi-quantitative assessment of STAS. By constructing multi-scale spatial topological maps from WSIs, we enriched the expression of histopathological image features. Notably, the TXY-CSU dataset achieved a substantial volume of 2758 FSs and PSs, representing the largest STAS dataset to date. This extensive collection enabled the DAEM model to thoroughly learn pathological features within WSIs, thereby enhancing STAS detection rates. Importantly, DAEM's capability to diagnose both FS and PS offers valuable guidance for intraoperative surgical decisions and postoperative adjuvant therapy plans. This process assists pathologists in identifying the presence of STAS in WSIs, improving diagnostic efficiency and accuracy while reducing instances of misdiagnosis and missed diagnoses. However, the DAEM model does exhibit certain false positives. On one hand, this may be due to the model learning reactive pathological changes commonly associated with tumors during the training process, and incorrectly identifying them as tumor morphological features. These changes, such as lymphocyte infiltration, hemorrhage, necrosis, stromal reactive hyperplasia, and fibrosis, often occur within the tumor region and are mixed with tumor cell nests. Pathologists also find it difficult to completely exclude these areas during manual annotation, making it easy for the model to learn them as tumor features. On the other hand, some normal tissue components may remain within the tumor region, such as carbon particle deposits, bronchial mucosa, muscular large blood vessels, and foam cells containing carbon particles within the airways. These tissue features bear a resemblance to tumor cells, and the model may mistake them for histological features of tumor cells, such as large deeply stained nuclei, elongated and twisted nuclear shapes, and individual or clustered tumor cells distributed in the airways, leading to misclassification and increased false positives. Conversely, the DAEM model also experiences false negatives, which can be attributed to certain sparse or individual STAS lesions that morphologically resemble foam cells containing carbon particles or nuclear fragments from necrotic material, causing the model to miss these lesions. In addition to these issues, the tissue preparation techniques also influence the model's diagnostic accuracy. Wrinkles, knife marks, and variations in staining reagents on the slides make image interpretation more challenging and contribute to both misdiagnosis and missed diagnoses by the model.

To investigate cross-center validation, we employed eight external cohorts to assess the generalizability and accuracy of the DAEM model in diagnosing STAS and non-STAS WSIs. Most external test sets demonstrated that DAEM achieved strong diagnostic performance for STAS. However, the model exhibited substantial variation in sensitivity (recall) and specificity across datasets, which directly affected the likelihood of false negatives and false positives. Specifically, in centers such as TXH-CSU, XH-CSU, PCPH, and CPTAC, precision exceeded recall, indicating that the model adopted a more conservative strategy when assigning STAS, thereby reducing false positives at the cost of missed detections. This imbalance suggests that subtle or small STAS lesions were under-recognized, likely due to weaker local context or site-specific staining variations that diluted discriminative features. In contrast, at TH-ZZU, FAH-NHU, and CJH, recall exceeded precision, reflecting improved sensitivity but accompanied by an increase in false positives. In these datasets, DAEM frequently misclassified macrophages, inflammatory cells, or stromal fibers as STAS, underscoring the risk of over-calling in morphologically complex microenvironments. Particularly concerning was the TCGA cohort, where overall accuracy appeared high, but the markedly lower F1 score revealed a skewed trade-off between false positives and false negatives, highlighting the pitfalls of relying solely on global metrics in class-imbalanced populations. Taken together, these findings emphasize that cross-center domain shifts manifest as divergent error patterns, and that robust multicenter deployment of DAEM will require site-specific calibration and adaptive normalization strategies to jointly suppress false positives while rescuing false negatives.

Interpretability is a critical dimension of model performance, as it enables clinicians to better understand the decision-making process and enhances confidence in algorithmic outputs. In this study, we interrogated DAEM through visualization of the TME density maps and multi-scale WSIs, thereby providing a comprehensive view of how the model arrives at its predictions. Both the density maps and the corresponding heatmaps closely mirrored the actual tumor locations, indicating that DAEM effectively prioritizes tumor cells and regions when assessing the presence of STAS. Patch-level analyses across different stages further confirmed that the model reliably distinguishes between high-, medium-, and low-contributing regions. These findings suggest that DAEM first learns discriminative features of tumor cells before incorporating signals from surrounding stromal or immune compartments. Notably, the interpretability of low-magnification patches was less consistent, whereas high-magnification patches yielded more faithful visualizations. Furthermore, visualizations derived from FS were inferior to those from PS, reflecting the morphological distortion introduced during freezing, where ice-crystal formation compromises nuclear and stromal integrity and thus hinders feature extraction and representation. Importantly, we extended interpretability analyses to external multicenter datasets, generating multi-scale heatmaps of STAS and non-STAS WSIs. For non-STAS cases, regions of interest highlighted by DAEM at 10× magnification closely resembled pathologists' annotations, capturing mesoscopic cues such as tumor bulk, tumor–stroma interfaces, and the continuity of alveolar structures. A 512×512 patch at 10× affords sufficient context to delineate glandular or solid blocks within their architectural surroundings, producing heatmaps well aligned with expert annotations. In contrast, 20× magnification proved more faithful in STAS cases, where the diagnostic hallmarks are inherently microscopic, which floating micropapillary clusters, detached tumor nests, and scattered single cells within alveolar spaces. At 10×, these subtle lesions occupy only a small fraction of each patch and are diluted by background alveoli, making attention signals diffuse. By contrast, 20× magnification ensures that the same-sized 256×256 patches amplify the relative proportion of STAS foci, enabling clearer recognition of cytologic details such as nuclear atypia, diminished cohesion, and luminal floating margins. Consequently, DAEM's attention maps at 20× are more congruent with the fine-grained regions of interest delineated by pathologists. Together, these results underscore that the interpretability of DAEM is inherently scale-dependent, reflecting the distinct structural versus cytological cues that dominate the recognition of non-STAS and STAS, respectively.

Here, we present an online STAS diagnosis and distance-measurement tool to assist pathologists with automated detection and semi-automated estimation of the distance from STAS foci to the primary tumor. First, DAEM-assisted STAS assessment improves interobserver consistency and reduces subjectivity, thereby lowering the risk of missed detections or under-reporting. Second, in resource-limited settings with heavy pathology workloads, this system could serve as an aid in secondary review or pre-screening, highlighting suspicious STAS areas, thereby improving efficiency and shortening turnaround time. For hospitals that currently lack dedicated thoracic pathologists or have not yet incorporated STAS into routine reports, the tool can function as part of a centralized pathology quality-control platform, enabling a "central second-opinion" mechanism and facilitating wider adoption of STAS reporting. Importantly, pathologists can leverage TME information on WSIs for cell typing and visually examine the spatial distribution of different cell populations. However, the accuracy of the nuclear segmentation model used in this tool still falls short of 100%, resulting in suboptimal cell type annotation within the TME. Consequently, this tool requires interactive usage with a pathologist to estimate the distance from STAS to the main tumor. Only through a human–machine interactive process can pathologists accurately pinpoint the distance between STAS and the primary tumor. We hope that in the future, the distance from STAS dissemination foci to the primary tumor can be directly obtained via the WSI-based TME, though this will necessitate continuous improvements in the accuracy of the lung cancer nuclear segmentation model. Secondly, the use of 3D tissue-based spatial modeling to measure the distance between STAS and the primary tumor still requires further investigation. Therefore, through broader involvement of clinical pathologists and thoracic surgeons, our STAS measurement tool will be further improved.

Unlike other related studies, we segmented and classified all cells within the WSIs and quantified relevant indicators in the TME. Through independent t-tests between WSIs with and without STAS, we found that the stromal-to-tumor ratio (STR), immune-to-tumor ratio (ITR), and MVD were statistically significant across the TCGA_LUAD, CPTAC_LUAD, and SXY-CSU datasets. Subsequently, we investigated TME differences between micropapillary-type STAS and non-micropapillary-type STAS. Using the Kruskal-Wallis test to analyze quantitative TME indicators, we discovered that MVD and stromal volume ratio (SVR) exhibited statistical differences in the TCGA, CPTAC, TXY-CSU, and SXY-CSU datasets. We speculate that STR, ITR, and MVD are important biomarkers for STAS, while MVD and SVR are crucial markers for micropapillary-type STAS. Additionally, we analyzed 16 quantitative indicators in the TME across different pathological types of STAS. As shown in Supplementary Table 4, the proportions of tumor cells and stromal cells in tumors containing micropapillary-type STAS differed significantly. When multiple STAS types were mixed, the tumor-to-stroma ratio changed notably. These TME-based analyses contribute to the understanding and research of STAS mechanisms, laying the foundation for precision medicine.

Despite significant advancements in the study of STAS, several limitations in the current experiments warrant attention. Although the DAEM model demonstrated robust performance on datasets from certain hospitals, its performance showed a slight decline when tested on datasets from other hospitals. Specifically, DAEM achieved high diagnostic accuracy on the XH-CSU and TXH-CSU datasets. However, its performance was suboptimal in hospitals with lower medical standards, such as CJH and PCPH. Additionally, the model's performance on the TCGA_LUAD and CPTAC_LUAD datasets was relatively lower. Published reports have noted the poor quality of WSI in these datasets, which may be attributed to issues with the scanning equipment leading to substandard image quality [82–84]. In summary, the variability in model performance can be attributed to multiple factors, including discrepancies in imaging devices across hospitals, which can result in inconsistencies in image quality and feature representation; biases in data collection, such as inconsistencies in labeling and sample imbalance; and environmental factors, such as disparities in hospital resources and staff training, all of which may impact the model's generalizability and practical application. To address these issues, we recommend performing fully automated quality control of WSIs prior to their use, ensuring the exclusion of those with quality issues. Tools such as HistoQC, PathProfiler, and HistoROI may aid in this quality control process [83–85]. Moreover, during data preprocessing and feature normalization, we propose the use of standardization algorithms such as Reinhard, Macenko, and Vahadane, along with GAN models, to standardize patches segmented from WSIs. Another significant challenge observed in this study is the issue of sample imbalance, which may have contributed to the suboptimal external validation performance of the DAEM model. In future work, we plan to address this by adding corresponding samples to the external validation datasets based on their specific characteristics. For pathology images, we will apply data augmentation strategies, including rotation, flipping, color jittering, and random cropping, and use GANs to generate realistic synthetic positive examples to mitigate the scarcity of positive samples. Similarly, undersampling techniques will be considered to balance the ratio of positive and negative examples in the training set. Furthermore, we will use cost-sensitive learning or Focal Loss to address sample imbalance and fine-tune the model on datasets from different hospitals to better adapt to specific environments and enhance generalization. In addition, we propose establishing a monitoring system for real-time feedback and model updates once deployed in clinical settings. Lastly, while the AUC for HD-Yolo segmentation of lung cancer WSI cells is approximately 0.85, there remain significant errors in our TME analysis. Future work should focus on developing more accurate models for identifying a wider range of cells related to the TME in WSIs. Currently, STAS localization remains semi-automated and does not yet meet the standard for full automation. Future studies should explore advanced models capable of automatically localizing STAS dissemination foci in WSIs. In conclusion, the recommendations outlined here aim to address challenges posed by variations in medical standards across hospitals, discrepancies in scanning equipment, and performance issues related to dataset quality. These efforts are crucial for improving the robustness and applicability of the DAEM model across diverse clinical settings.

In summary, we propose a DAEM for diagnosing the presence or absence of STAS in lung cancer FS and PS. The diffusion attention expert module in DAEM leverages full-attention aggregation based on a diffusion attention mechanism to learn multi-scale features in histopathology images, and its dual-branch structure facilitates the learning of multi-scale feature representations. In our internal dataset, DAEM achieves clinical-grade performance in diagnosing STAS on both FS and PS, providing a reliable tool for rapid intraoperative STAS diagnosis and large-scale postoperative screening. Importantly, we validate the generalizability and interpretability of DAEM on STAS test datasets from eight additional centers. Furthermore, for the first time, we employ the distribution of cell nuclei in the TME to semi-automatically measure the distance between STAS and the primary tumor. Notably, to explore the significance of quantitative indicators in the TME, we analyze datasets from multiple centers and find that the STR, the ITR, and MVD serve as biomarkers for STAS histopathology images, while MVD and the SVR are biomarkers for histopathology images containing micro-lobular STAS. Therefore, our study introduces a DAEM method and a STAS distance measurement tool for diagnosing and analyzing STAS in histopathology images, offering a valuable reference and research direction for precision oncology.

## Methods

This study was conducted in accordance with the Declaration of Helsinki. Model development and internal validation were performed using retrospective, anonymized WSIs from patients at the Second Xiangya Hospital of Central South University (SXH-CSU). The study protocol for these internal data was approved by the Ethics Committee of SXH-CSU (Ethics Review Document Z0331-01). External validation was conducted using anonymized, retrospective WSIs from eight additional centers. As all external datasets contained no patient-identifiable information and only de-identified, retrospective WSIs were used for model evaluation, no additional ethics approval was required for these datasets in accordance with local regulations and institutional guidelines.

### Multicenter dataset collection

In this retrospective, multi-center study, we utilized anonymized hematoxylin and eosin (H&E) stained lung cancer slides from six hospitals and two projects in China and the United States, constructing nine cohorts for model training and validation. Based on our research objectives, only patients meeting the following criteria were included: (1) diagnosed with lung adenocarcinoma; (2) corresponding routine pathological slices, including primary tumor tissue and adjacent non-tumor tissue; (3) detailed TNM staging; (4) high-quality slides, such as those without bending, wrinkling, blurring, or color changes; (5) absence of tumor cells randomly distributed, with irregular pericellular nests typically located at the tissue slice edge or outside the tissue slice; (6) absence of tumor cell continuity spreading from the tumor edge to the most distant airway; (7) absence of benign cytological features of lung epithelial cells or bronchial cells and/or presence of cilia; (8) absence of linear cellular strips detached from the alveolar wall in histopathological images. Given that pathological diagnosis of STAS is prone to misdiagnosis and missed diagnosis, three experienced pathologists independently labeled each WSI for STAS under a double-blind principle with cross-validation, ensuring accuracy and reducing subjectivity, underdiagnosis, or overdiagnosis. During the inclusion and exclusion process, WSIs with mechanical artifacts had already been removed by these pathologists with many years of diagnostic experience. Although false positives in STAS identification may indeed occur in clinical pathology practice, we employed relevant immunomarkers (TTF-1, CK, CD68) to stain suspicious lesions, thereby further distinguishing true STAS-positive dissemination from histiocytes (Supplementary Figure 7). We included WSIs from the cohort of the Second Xiangya Hospital of Central South University (SXH-CSU) as internal training and validation data. This cohort selected 594 patients diagnosed with STAS and 315 patients without STAS from 12169 patients who underwent lung nodule resection at the Second Xiangya Hospital between April 2020 and January 2024. The experiment collected 2758 WSIs (including 526 FSs and 2232 PSs, approximately 1-4 WSIs are retained for each patient), immunohistochemical image data, and related clinical information from the selected patients (Supplementary Table 5).

The external validation set includes pathological images from eight centers. The cohort from the Third Xiangya Hospital of Central South University (TXH-CSU) provided 304 slides from 68 patients diagnosed with STAS between 2022 and 2023. The cohort from Xiangya hospital of Central South University (XH-CSU) provided 214 WSIs from 190 patients diagnosed with STAS between 2022 and 2023. The cohort from the Affiliated Tumor Hospital of Zhengzhou University (TH-ZZU) provided 91 WSIs from 19 patients diagnosed with LUAD and STAS in 2023. The cohort from the First Affiliated Hospital of Nanhua University (FAH-NHU) selected 130 WSIs from 42 patients diagnosed with LUAD and STAS between 2021 and 2024. The cohort from Changsha Jingkai Hospital (CJH) selected 91 WSIs from 45 patients diagnosed with LUAD and STAS between 2023 and 2024. The cohort from Pingjiang County First People's Hospital (PCPH) selected 78 WSIs from 35 patients diagnosed with LUAD and STAS between 2019 and 2021. The TCGA_LUAD dataset includes 366 patients with 417 WSIs. The CPTAC_LUAD cohort includes 170 patients with 443 WSIs. In the validation cohorts, TCGA_LUAD and CPTAC_LUAD provided patient-related clinical and multi-omics data, offering valuable information for survival and mechanistic studies of STAS patients. For detailed statistical data, please refer to the appendix (Supplementary Table 6).

**Model development process**

The STAS diagnosis and localization process we proposed mainly consists of three steps: (1) image preprocessing (Figure 1a, 1b, 1c, 1d) (2) image feature preprocessing, feature extraction and diagnosis. (3) Multi-center validation and interpretability analysis. Among them, image preprocessing mainly includes the extraction of lung cancer tissue and the digitization of WSI. The Otsu algorithm divides the WSI into tissue, background and blurred areas. Then, the sliding window strategy is used to divide the 20-based WSI into $256 \times 256$ and $512 \times 512$ patches, respectively, and the spatial coordinates of the patches are obtained. To mitigate the influence of variability in scanning devices and slide quality on the model's generalization performance, all patches across datasets were preprocessed and standardized using the Macenko normalization algorithm prior to feature extraction with CTransPath [54,54,86]. Given that the pathological diagnosis of STAS is prone to misdiagnosis and missed diagnosis, we use immunohistochemistry and triple cross-validation methods to obtain the diagnostic results of WSI. The DAEM framework was trained to construct the diagnostic STAS model by leveraging both the features and labels of all patches within the WSIs. Subsequently, multi-center external datasets, following image and feature preprocessing, were employed for prediction and validation to evaluate the generalizability of DAEM. In addition, interpretability analyses were conducted to elucidate the decision-making process of DAEM.

**Feature preprocessing**

Given that pathologists review $20\times$-based WSIs, they analyze and diagnose based on multi-scale features, including global pathological features at low magnification, tissue features at medium magnification, and cell features at high magnification. Therefore, in our experiments, we employ a multi-scale feature extraction method to extract pyramid-structured features from WSIs. To efficiently capture multi-scale local semantic information in the images and enhance the performance of subsequent tasks, we use CTransPath [87] as the feature extractor to obtain patch features of sizes $256 \times 256$ and $512 \times 512$. CTransPath is a pre-trained model based on the Swim transformer structure, which is used to uniformly represent all patches as 768-dimensional one-dimensional vectors [88]. After feature preprocessing, each WSI can be represented as a set of one-dimensional features. However, there exist spatial semantic relationships among the patches in the WSI. Therefore, based on the spatial positional relationships of the patches, we construct a spatial topological graph ($G_l, G_s$) of the WSI using the KNN ($K = 9$) nearest neighbor algorithm, where nodes represent the patch features and edges represent the relationships between patches [89]. Secondly, in order to construct the TME features in the WSI, we use the HD-Yolo to segment and classify the cells in the WSI, and construct the nodes and edges of the spatial topology graph ($G_{TME}$) of the TME based on the spatial position relationship of the cells [42]. Nodes represent cell labels, classification probabilities, and area-related features of cell nuclei, and edges represent the relationships between cells of the same species. Therefore, WSI can be expressed as $G_l$, $G_s$, $G_{TME}$, where $G_l = (V_l, E_l)$, $G_s = (V_s, E_s)$, $G_{TME} = (V_{TME}, E_{TME})$. The data preprocessing procedures for the internal training and validation sets as well as the external validation set were consistent, including patch segmentation, feature extraction, and the construction of spatial topological graphs for both WSIs and the TME.

**Multi-scale feature representation**

In the process of feature extraction, we proposed a diffusion attention expert model to extract effective feature representation from multi-scale features. Diffusion Attention Expert Model (DAEM) includes a multi-scale graph convolution module (MSGC) and two expert diffuse attention modules. Among them, MSGC is used to process the multi-scale spatial topology of histopathological images. MSGC is mainly composed of two sets of three parallel SAGEConv layers, LeakyReLU layer, LayerNorm layer and Dropout layer. Given that the graph constructed by multi-scale features is large, MGCM uses SAGEConv to extract features in $G_l$, $G_s$, $G_{TME}$, [90]. This is because SGAEConv uses attention weighting to aggregate neighborhood information, not only considering local geometric structure, but also combining global features to improve feature expression capabilities. Specifically, the network consists of two steps, namely fusing neighbor node features and updating current node features. The above process is as follows:

$$\begin{aligned} AGG &= \sum_{u \in N(v)} h_u^{(l)} \Big/ |N(v)| \\ h_v^{(l)} &= \sigma(W \square AGG(\{h_v^{(t)}\} \cup \{h_u^{(l)}, \forall u \in N(v)\})) \end{aligned} \tag{1}$$

Where $h_v^{(l)}$ represents the feature vector of node $v$ in layer $l$, $N(v)$ represents the neighbor set of node $v$, $|.|$ indicates the number of elements in the collection. $W$ and $\sigma$ are trainable weights and LeakyReLU functions, respectively [91]. Secondly, in order to alleviate the problems of gradient vanishing and neuron death, MGCM uses the activation function LeakyReLU instead of ReLU. In order to normalize all features after aggregation and make the vector representation of features within a stable numerical range, MGCM uses LayerNorm to accelerate model training and improve the generalization performance of the model. Finally, MGCM uses Dropout to alleviate the overfitting of graph neural networks. The final feature extraction process can be simplified as:

$$h_v^{l+1} = Dropout(LayerNorm(Leaky\,\mathrm{Re}\,LU(h_v^t))) \tag{2}$$

**Diffusion attention expert module**

The process of further feature extraction is primarily accomplished by two independent diffusion attention expert modules (DAEM). Each DAEM consists of three adaptive pooling layers, one feature aggregation layer, and a diffusion attention module (DAM). The multi-scale features processed by MSGC are represented as three one-dimensional features of different lengths. Then, these features of varying lengths contain a large amount of redundant information, so the model needs to reduce its focus on the redundant information while enhancing its attention to diagnostically important information to improve its generalization ability. The expert diffusion attention module uses AdaptiveMaxPool1d (AMP1d) to map features of different lengths to fixed-length features, thereby extracting the key information and reducing computational complexity.

Given that the lengths of the one-dimensional representations of multi-scale features of WSI vary greatly,

AdaptiveMaxPool1d pools the one-dimensional features of $10\times$, $20\times$, and TME into one-dimensional feature representations of lengths of 512, 2048, and 2048, respectively. A large number of pre-experiments have also proven that this fixed length is the best parameter for this module. The concatenated feature $v$ are then passed to the DAM for feature extraction. The DAM is a neural network module designed for information transfer by simulating the diffusion process, inspired by the thermodynamic process of heat conduction. In this process, each node in the system represents a sample, each with an initial state (such as temperature), and there is a flow of signals between the nodes. Over time, the states of these nodes are continuously updated. In the context of neural networks, these state changes can be seen as the forward propagation process of sample embeddings, where interactions between samples are updated through information transfer. A classic diffusion process can be described by a heat conduction equation (a partial differential equation with initial conditions):

$$\frac{\partial Z(t)}{\partial t} = \nabla^* \left( S(Z(t),t) \circ \nabla Z(t) \right), \quad \text{subject to} \quad Z(0) = \left[ x_i \right]_{i=1}^{N}, \quad t \geq 0 \tag{3}$$

Here, $\nabla$, $\nabla^*$ and $S(Z(t),t)$ represent the gradient operator, divergence operator, and diffusivity, respectively. Therefore, in the description of the diffusion process, the classical heat conduction equation is used to describe the flow and changes of signals between nodes. In a discretized space, the gradient, divergence, and diffusivity correspond to the signal differences between nodes, the total amount of signal flowing out of a node, and the rate of signal flow between nodes, respectively. This process can be discretized and iteratively updated using numerical methods, thereby describing the change in node states at each moment in time:

$$\frac{\partial z_i(t)}{\partial t} = \sum_{j=1}^{N} S_{ij}(Z(t),t) \left( z_j(t) - z_i(t) \right) \tag{4}$$

Here, $S_{ij}(Z(t),t)$ is the diffusion rate between nodes $i$ and $j$, while $z_j(t)$ and $z_i(t)$ represent the states of nodes $i$ and $j$, respectively. In the DAM, the diffusion process is used to update node representations through inter-layer update mechanisms. By introducing numerical iteration (such as the explicit Euler method), the continuous diffusion process is transformed into discrete update steps. Specifically, the update formula for node representations is:

$$z^{(k+1)} = S \cdot z^{(k)} + \mathrm{Re}\,sidual \tag{5}$$

Here, $k$ represents the current layer, $D$ is the diffusion matrix, denotes the influence between nodes, and *Residual* is the residual connection from the previous layer's representation. To optimize the update process of node embeddings, an energy function is introduced to describe the inherent consistency between nodes. The energy function is formulated as:

$$E = \alpha \left\| z_i - z_i^{'} \right\|^2 + \beta \sum_{i,j} \left\| z_i - z_j^{'} \right\|^2 \tag{6}$$

The first term constrains the local consistency of each node with respect to itself, while the second term enforces global consistency between the nodes. By minimizing the energy function, the node embeddings are guided towards an optimal state. The DAM controls the influence between nodes at each layer through an adaptive diffusion rate matrix. The representation update at each layer not only depends on the node's own state but also considers the embeddings of all other nodes, thereby enhancing the model's expressive power. This update mechanism is similar to the multi-head attention in Transformers, but by introducing the concept of diffusion rate, the model can dynamically adjust the influence between different nodes. Specifically, The diffuse attention module first linearly maps $v$ to obtain query ($Q$), key ($K$), and value ($v$) vectors, respectively. The process is as follows:

$$Q = W_Q v; K = W_K v; V = W_V v,$$
$$Q \in R^{N \times H \times M}; K \in R^{L \times H \times M}; V \in R^{L \times H \times D}, \tag{7}$$

The shape of the $Q$ tensor is $[N, H, M]$, where $N$ represents the number of queries, $H$ represents the number of attention heads, and $M$ represents the feature dimension of each query (expanded as a dimension in the attention calculation). The shape of the $K$ tensor is $[L, H, M]$, where $L$ represents the number of keys. The shape of the $V$ tensor is $[L, H, D]$, where $D$ is the dimension of the $V$ vector. Then, $Q$ and $K$ are normalized to get:

$$\tilde{Q} = \frac{Q}{\left\|\tilde{Q}\right\|_2}; \tilde{K} = \frac{K}{\left\|\tilde{K}\right\|_2} \tag{8}$$

The molecular part of attention aggregation is to first calculate an intermediate tensor $KV_{H,M,D}$, and then calculate the attention molecule. The process is:

$$KV_{H,M,D} = \sum_{l=1}^{L} \tilde{K}_{L,H,M},$$
$$num(N,H,D) = \sum_{l}^{L} \tilde{K}_{N,H,M} KV_{H,M,D} + \sum_{l=1}^{L} V_{L,H,D} \tag{9}$$

Then use $Q$ to calculate the normalization factor and include a bias in the normalization factor. The process is:

$$denom(N,H) = (\sum_{m=1}^{M} \tilde{Q}_{N,H,M} (\sum_{l=1}^{L} \tilde{K}_{L,H,M})) + bias \tag{10}$$

The final output attention result can be obtained by dividing the numerator by the normalization factor, and the process is:

$$y_{out}(N,H,D) = \frac{1}{N} \sum_{h=1}^{H} \frac{num(N,H,D)}{denom(N,H)} \tag{11}$$

After the two expert modules have completed feature extraction, the classification module integrates the features provided by the expert models from different branches for final classification. The classification module is mainly composed of the Concatenation layer, the Layer Normalization layer, the Linear layer, and the Dropout layer [92]. The classification module uses the Concatenation layer to concatenate the feature representations output by the two expert models, and then uses the Layer Normalization layer to normalize the feature representations. The classification module uses the Linear layer to reduce the dimension of the features. The Dropout layer randomly sets the output of some neurons to zero to prevent overfitting. Finally, the entire model is classified through the Linear layer. The whole process is as follows:

$$y = W_2 \cdot Dropout(W_1 \cdot LayerNorm([x_1, x_2] + b_1) + b_2) \tag{12}$$

Among them, $x_1$ and $x_2$ represent the output feature representations of expert model 1 and expert model 2, respectively. $W_1 \in R^{64\times256}$, $W_2 \in R^{2\times256}$, $b_1 \in R^{64}$, $b_2 \in R^2$ are the weight matrix and bias respectively.

**Training strategy**

Assuming that different expert models have the same understanding of the same feature, the outputs of the two expert models should be the same during model training. We propose a supervised contrastive learning loss and mean square error loss to constrain the expert diffusion module to extract feature representations that are relevant and consistent with the results. Given a multi-scale feature $z_i$ extracted by the WSI model, its category label is $y_i$. For each multi-scale feature $z_i$, its supervised contrast loss is defined as follows:

$$SupConLoss = \sum_{i\in I} \frac{-1}{|P(i)|} \sum_{p\in P(i)} \log \frac{\exp(z_i \cdot z_p / \tau)}{\sum_{a\in A(i)} \exp(z_i \cdot z_a / \tau)} \tag{13}$$

$\tau$ represents the temperature scalar to control the smoothness of the distribution. $A(i)$ represents the set of all sample indices except itself. The positive sample set $P(i)$ is the index of all samples with the same category label as sample $i$ (excluding $i$ itself). Assuming the predicted value and the true value of the given model, the mean square error (MSE) loss is calculated as follows:

$$MSELoss = \frac{1}{n} \sum_{i=1}^{n} (y_i - \hat{y}_i)^2 \tag{14}$$

Where $n$ represents the total number of samples. For the final classification features, the model uses the cross entropy loss function to optimize the performance of the model, and the process is:

$$CELoss = -\frac{1}{N} \sum_{i=1}^{N} \sum_{j=1}^{C} y_{ij} \log(\hat{y}_{ij}) \tag{15}$$

Therefore, the process of model optimization can be summarized as reducing the loss of supervised contrastive learning, the sum of mean square error loss and cross entropy loss. We assign different weights to different loss functions, which can be expressed as:

$$\min_{\theta} Loss_{all} = \min_{\theta} \sum_{i=1}^{N} \lambda_i SupConLoss + \beta_i MSELoss + \gamma_i CELoss \tag{16}$$

Among them, $\lambda \in (0,1)$, $\beta \in (0,1)$ and $\gamma \in (0,1)$ are hyperparameters respectively. Through these hyperparameters, we can dynamically adjust the impact of different losses on the overall optimization process.

**Baseline Models**

CLAM_MB: The CLAM framework is based on attention mechanism-based MIL, improving model performance and data efficiency through instance-level clustering and attention pooling. It also utilizes an attention network to predict unique attention scores for each class, further refining the feature space [56]. CLAM_MB uses a single-branch attention mechanism, generating attention scores only for a single class (such as the positive/negative class in binary classification problems) [56].

CLAM_SB: It adopts a multi-branch attention mechanism, designing independent attention branches for each class. The goal is to improve the model's ability to distinguish complex pathological features by learning the class-specific feature weights while handling multi-class tasks [56].

ABMIL: This is an improved method within the MIL framework, which frames the MIL problem as learning Bernoulli distributions for bag labels. The core idea is to automatically learn the importance weights of different instances in the bag through the attention mechanism [57].

DSMIL: It uses a trainable distance metric to model the relationships between instances in a dual-stream architecture [45]. To address the issue of large or imbalanced bags in MIL model training, DSMIL employs self-supervised contrastive learning to extract robust patch representations and then utilizes a pyramid fusion mechanism to integrate multi-scale WSI features [45].

IBML: Intervention Bag Multi-Instance Learning (IBMIL) achieves intervention training through backdoor adjustment, thereby suppressing bias caused by prior bag context and enabling deconfounded bag-level predictions [58].

ILRA: Leveraging the low-rank structure within the high-resolution WSI data manifold, we enhance feature embedding through contrastive learning with a Low-Rank Constraint (LRC) and introduce an iterative low-rank attention mechanism during the feature aggregation phase to simulate global interactions among all instances, thereby improving the performance of gigapixel-level WSI classification. [63].

ACMIL: Mitigates overfitting by employing multiple branch attention and stochastic top K instance masking to reduce attention value concentration and capture more discriminative instances in WSI classification [59].

DGRMIL: Models instance diversity by converting instance embeddings into similarities with predefined global vectors via a cross-attention mechanism and further enhances the diversity among these global vectors through positive instance alignment and a determinant point process-based diversified learning paradigm [61].

TransMIL: It is a transformer-based MIL framework that leverages both morphological and spatial correlations among instances to enhance visualization, interpretability, and performance in WSI pathology classification [49].

DTFD-MIL: Addresses the challenge of limited WSI samples in MIL by introducing pseudo-bags to virtually enlarge the bag count and implementing a double-tier framework that leverages an attention-based derivation of instance probabilities to effectively utilize intrinsic features [52].

PathGCN: It is a method for learning spatial operators from random paths on a graph, aiming to avoid over-smoothing by combining learned spatial operators with pointwise convolutions, thereby enhancing the expressiveness and performance of Graph Convolutional Networks (GCN) [60]。

MIHM: It is a multi-instance learning framework that takes image patches from pathological slides as basic instances and models and aggregates these local features to obtain a comprehensive representation of the WSI. Unlike traditional methods, MIHM not only considers the contribution of individual patches but also leverages a hierarchical structure to capture multi-level spatial relationships and semantic information, thereby enhancing the model's capacity to represent complex pathological features and improving diagnostic accuracy[62].

**Experimental setup and implementation details**

The model evaluates its performance using five-fold cross-validation based on the WSIs in the SXH-CSU dataset. In this cross-validation method, the frozen and paraffin-embedded sections in the SXH-CSU dataset are evenly divided into five groups. To avoid homologous sample leakage, all WSIs were grouped according to their patient IDs, ensuring that slices from the same patient were not split across subsets. Each time, four groups are used for training, and the remaining one group is used for validation, resulting in five optimal prediction models. Then the results of DAEM's five-fold cross validation on the SXH-CSU dataset are counted (Supplementary Table 7). All baseline models and DAEM are trained and evaluated on the same folds. Most importantly, the model's generalization ability and clinical applicability are validated using an external, independent dataset from eight centers.

We performed a pre-experiment to grid search the optimal learning rate, regularization strength, and batch size, ultimately determining the current parameter combination. The DAEM is trained using the AdamW optimizer with a learning rate of 0.001, and the learning rate is adaptively adjusted during the model training process. The batch size is set to 1 primarily because the number of patches in each WSI varies, and the size of the graph adjacency matrix also differs. Thus, it is impossible to stack graphs of different sizes into a batch with a unified output. Furthermore, the model consumes a large amount of GPU memory during inference, and increasing the batch size could lead to memory overflow. The number of epochs is set to 100, with one iteration over the training data per epoch. For comparison experiments, we implemented state-of-the-art methods for CLAM_SB, CLAM_MB, ABMIL, DSMIL, DTFD, IBMIL, ACMIL, PatchGCN, DGRMIL, TransMIL, MIHM, ILRA, and DAEM. The Dropout value in the model is 0.2.

**Visualization and explainability**

To generate attention-based heatmaps for visualizing WSIs, we implemented a multi-step process that integrates patch-level contributions with spatial localization. Initially, we partition the WSI into 256×256 and 512×512 patches using OpenSlide or similar tools. Each patch undergoes color normalization to standardize visual features. Subsequently, we employ a pre-trained model, such as CTransPath, to extract feature vectors for each patch. These vectors are input into the DAEM model, which assigns an attention weight to each patch. To mitigate potential biases in contribution, we average the contributions across symmetric mixed encoder outputs and normalize them to a [0,1] range. These normalized contributions are then mapped back onto the original WSI using their respective spatial coordinates. For visualization, we utilize matplotlib or the built-in visualization scripts provided by the tools, generating colorful heatmaps that overlay on grayscale or pseudocolored slices, facilitating observation and analysis by pathologists. The specific code used to generate these heatmaps is available at https://github.com/panliangrui/DAEM.

To validate the model's interpretability and accuracy, we visualized and analyzed both global and local areas of interest in FS and PS slides containing STAS and non-STAS regions, comparing the areas highlighted by both pathologists and the DAEM model. Additionally, in external testing, we applied this visualization technique to assist users in understanding the decision-making process of DAEM across different centers. Finally, based on the patch contribution scores output by DAEM, we identified the top 4 patches from high, medium, and low contribution levels for further analysis.

**Hardware and software**

Data processing was performed on a Windows 11 host with an Intel i7 13700 K processor, 128GB of memory, 100TB of storage, and a 24GB NVIDIA GeForce RTX 4090 graphics card. Model development, testing, and training were performed using PyTorch 1.13, Python 3.9, and various other open-source tools and packages.

## Statistics & Reproducibility

We used ROC curve, PRC curve, calibration curve, accuracy, precision, recall, F1 Score, specificity, and AUC as evaluation metrics for model performance. The ROC curve represents the relationship between sensitivity and specificity, with a larger area indicating better performance. The PRC curve shows the relationship between precision and recall, with a larger area indicating better performance. Higher values for accuracy, precision, recall, F1 Score, specificity, and AUC indicate better performance. In the five-fold cross validation, we used the DeLong test to compare the AUC values and other evaluation indicators of the baseline model and the DAEM model [93]. In addition to the internal dataset (SXH-CSU), the model was tested on independent datasets from eight external institutions to verify its generalization ability. Sample sizes were determined based on the availability of eligible cases in each dataset and were comparable to those used in previous studies; no formal statistical method was used to predetermine sample size. No data were excluded from the analysis unless they were of insufficient quality (e.g., severe artifacts or incomplete annotations), and all exclusion criteria were applied consistently across datasets. The experiments were not randomized. The investigators were not blinded to group allocation during experiments and outcome assessment. All quantitative results are reported as mean ± SEM. Statistical comparisons between two groups were performed using two-sided independent t-tests. Comparisons among three or more groups were performed using the Kruskal-Wallis test followed by Dunn's multiple comparison test. Survival differences between high- and low-expression groups were assessed using the Kaplan–Meier method with log-rank test. P values $< 0.05$ were considered statistically significant. Sample sizes (n) indicate the number of independent patients, with no pooling of technical replicates.

## Data availability

Due to privacy and ethical constraints related to multicenter data, the dataset upon which this study's results are based is not currently publicly available. You can contact the corresponding author to request data access, but approval is required. Specific access conditions include ethical review and data usage agreements. Data requests should be sent to the corresponding author's email address, and we will review each case individually. We expect to respond to data access requests within 4-6 weeks. Once approved, the data will be made available for access within six months, and access will only be permitted to eligible individuals or institutions for academic research purposes. We provide the labels and features of TCGA_LUAD and CPTAC_LUAD datasets at https://github.com/panliangrui/DAEM or https://zenodo.org/records/11611418. TCGA_LUAD and CPTAC_LUAD diagnostic pathology data can be downloaded at https://portal.gdc.cancer.gov/.

## Code availability

The source code and standalone program of DAEM are available at https://github.com/panliangrui/DAEM or https://zenodo.org/records/18824683. All code is for academic use only.

## Acknowledgements

This work was supported by National Key R&D Program of China 2023YFC3503400, 2022YFC3400400; NSFC-FDCT Grants 62361166662; The Innovative Research Group Project of Hunan Province 2024JJ1002; Key R&D Program of Hunan Province 2023GK2004, 2023SK2059, 2023SK2060; Top 10 Technical Key Project in Hunan Province 2023GK1010; Key Technologies R&D Program of Guangdong Province (2023B1111030004 to FFH); Hunan Province Graduate Research Innovation Project CX20240450; National Natural Science Foundation of China Youth Science Fund Project: 82200019. The Funds of State Key Laboratory of Chemo and Biosensing, the National Supercomputing

Center in Changsha (http://nscc.hnu.edu.cn/), and Peng Cheng Lab, The Hunan Province Graduate Research Innovation Project CX20240450. Peng Cheng Lab. Q. Liang and S.L. Peng are the corresponding authors for this paper.

## Author Contributions Statement

Liangrui Pan: Conceptualization, Methodology, Software, Visualization, Writing original draft, data collection, Writing-review & editing,. Jiadi Luo: Conceptualization, Writing original draft. Yuxuan Xiao, Chenchen, Songqing, Ling Chu, Li Manqiu, Xiaoshuai Wu, Rongfang He, Zhenyu Zhao, Ruixing Wang, Shulin Liu, Yiyi Liang: data collection. Qingchun Liang: Visualization, Writing original draft, Supervision. Xiang Wang and Shaoliang Peng: Funding acquisition, Resources, Writing-review & editing, Supervision.

## Competing Interests Statement

The authors declare that they have no known competing financial interests or personal relationships that could have appeared to influence the work reported in this paper.

## Figure Legends

Figure 1. Workflow of collection and organization of lung cancer STAS dataset, model training and inference, and multi-center validation. **a**, Resection of lung tumor tissue. **b**, Digitization of FSs and PSs. **c**, Histopathology image slice processing, **d**, Pathologists' triple cross-validation to annotate data. **e**, Extraction of multi-scale histopathology image features and parallel expert models to diagnose and localize STAS in histopathology images. **f**, Details of the expert module, where AMP1d is the AdaptiveMaxPool1d layer. **g**, Details of the classifier. **h**, Distribution and quantity of multi-center data. Internal training and validation data were collected from SXH-CSU (594 STAS patients, 315 non-STAS patients, 2758 WSIs). External validation datasets included eight cohorts: TXH-CSU (304 WSIs, 68 patients), XH-CSU (214 WSIs, 190 patients), TH-ZZU (91 WSIs, 19 patients), FAH-NHU (130 WSIs, 42 patients), CJH (91 WSIs, 45 patients), PCPH (78 WSIs, 35 patients), TCGA_LUAD (417 WSIs, 366 patients), CPTAC_LUAD (443 WSIs, 170 patients).

Figure 2. Experimental results of DAEM in predicting STAS on the dataset from the SXH-CSU after training with five-fold cross-validation. **a**, ROC curves for DAEM diagnosing FSs across five folds of cross-validation. The five curves represent the diagnostic performance of DAEM after each fold, with the diagonal line showing the in-domain test results. **b**, PRC for DAEM diagnosing FSs across five-fold cross-validation. **c**, Accuracy, precision, recall, F1 score, specificity, and AUC values for FSs diagnosis obtained through five-fold cross-validation. **d**, Confusion matrix for DAEM diagnosing FSs using five-fold cross-validation. **e**, ROC curves for DAEM diagnosing PSs using five-fold cross-validation. **f**, PRC curves for DAEM diagnosing PSs using five-fold cross-validation. **g**, Accuracy, precision, recall, F1 score, specificity, and AUC values for PSs diagnosis obtained through five-fold cross-validation. **h**, Confusion matrix for DAEM diagnosing PSs using five-fold cross-validation. The SXH-CSU cohort included 594 STAS patients and 315 non-STAS patients, with 526 FSs and 2232 PSs (1–5 WSIs per patient).

Figure 3. MIL methods predict the SOTA results of STAS in FSs and PSs. **a**, SOTA results of MIL methods and DAEM for predicting FSs. **b**, SOTA results of MIL methods and DAEM for predicting PSs.

Figure 4. Evaluation of DAEM's performance in diagnosing STAS in WSIs in multi-center experiments. **a**, ROC and PRC curves validating DAEM's diagnostic performance based on the STAS test set from XH-CSU (214 WSIs, 190 patients). **b**, ROC and PRC curves validating DAEM's diagnostic performance based on the STAS test set from TXH-CSU (304 WSIs, 68 patients). **c**, ROC and PRC curves validating DAEM's diagnostic performance based on the STAS test set from TH-ZZU (91 WSIs, 19 patients). **d**, ROC and PRC curves validating DAEM's diagnostic performance based on the STAS test set from FAH-NHU (130 WSIs, 42 patients). **e**, ROC and PRC curves validating DAEM's diagnostic performance based on the STAS test set from CJH (91 WSIs, 45 patients). **f**, ROC and PRC curves validating DAEM's

diagnostic performance based on the STAS test set from PCPH (78 WSIs, 35 patients). **g**, ROC and PRC curves validating DAEM's diagnostic performance based on the STAS test set from TCGA_LUAD (417 WSIs, 366 patients). **h**, ROC and PRC curves validating DAEM's diagnostic performance based on the STAS test set from CPTAC_LUAD (443 WSIs, 170 patients). **i**, Accuracy, precision, recall, F1 score, specificity, and AUC values for DAEM's prediction of the presence or absence of STAS on eight external test sets. Each WSI represents an independent patient; 1–4 WSIs per patient were retained for evaluation.

Figure 5. A flowchart employing human–computer interaction to determine the maximum distance from STAS to the primary tumor, thereby guiding clinical surgery. **a**, The emphasis placed by pathologists on both the primary tumor and STAS dissemination foci, and the corresponding focus of HD-yolo on STAS-related tumor cells within the TME. **b**, An interactive STAS distance measurement website; the associated code can be downloaded at https://github.com/panliangrui/DAEM. Users can select two points to measure the distance, such as the perpendicular distance from the most distant STAS dissemination focus to the tangential line at the primary tumor margin.

Figure 6. Multi-scale features significantly enhance the interpretability of NAVF-Bio's performance. **a**, Pathologist-annotated WSI of FS, tumor density map, and attention heatmaps at 20× and 10× magnifications. The model's attention scores are mapped back to the WSI, with different color codes representing tumor activity in each region. The heatmaps use the matplotlib jet colormap to indicate contribution levels, with red representing high-risk regions and blue representing low-risk regions. **b**, Based on DAEM's attention scores, the top 4 patches with high, medium, and low attention contributions in FS are shown. **c**, Pathologist-annotated WSI of PS, tumor density map, and attention heatmaps at 20× and 10× magnifications. **d**, Based on DAEM's attention scores, the top 4 patches with high, medium, and low attention contributions in PS are shown.

Figure 7. Multicenter validation of the DAEM model using multi-scale heatmaps to diagnose the presence of STAS in WSIs. The WSIs of the eight externally validated centers (XH-CSU, TXH-CSU, TH-CSU, FAH-CSU, CJH, PCPH, TCGA_LUAD, CPTAC_LUAD) were first manually annotated by pathologists. Then, the pathological features of randomly selected STAS were displayed, and interpretability analysis was performed on the heatmaps at different scales in DAEM.

Figure 8. Analysis of potential STAS biomarkers based on quantitative TME characteristics. **a**. TME visualization of the STAS group and the non-STAS group. **b-d**. Box plots and t-test results for inter-group comparisons of potential TME biomarkers in the TCGA_LUAD, and SXY-CSU datasets. Each box plot is overlaid with a single sample point to show the data distribution. In the box plots: the center line represents the median, the boxes represent the interquartile range (IQR), the maximum/minimum values are indicated, and single points represent independent sample data. **e**. TME visualization of STAS micropapillary subtypes. **f-g**. Across the TXH-CSU (STAS with MC, n = 51; Non-MC with STAS, n = 183; non-STAS, n = 80), SXH-CSU (STAS with MC, n = 460; Non-MC with STAS, n = 724; non-STAS, n = 996), CPTAC_LUAD (STAS with MC, n = 123; Non-MC with STAS, n = 144; non-STAS, n = 174), and TCGA_LUAD (STAS with MC, n = 58; Non-MC with STAS, n = 95; non-STAS, n = 262) datasets, the Kruskal – Wallis test was applied to identify potential biomarkers (MVD and SVR) associated with micropapillary STAS, with individual data points and box plots displayed for each group. **h-i**. Survival and recurrence risk curves for STAS-positive patients based on the SXH-CSU, CPTAC_LUAD, and TCGA_LUAD datasets. Statistical notes (applicable to b-d, f-g): Plots b-d used a two-sided independent samples t-test to assess differences in variables between the STAS and non-STAS groups; plots f-g used a two-sided Kruskal-Wallis test to assess differences across groups; significance is expressed as a p-value (as shown in the figure), and the significance sign is defined in the figure caption; the sample size (n) for each group is indicated in the figure, and all data represent biological replicates (each sample is from a different patient, not a technical replicate). Data are from unported tumor sections (WSI) of independent subjects, with each unit representing an independent patient sample; there was no technical replicate pooling between different groups; a single tumor divided into multiple sampling sites is not considered an independent replicate. Data are presented as mean values +/- SEM" as appropriate.

Figure 9. Workflow for DAEM-assisted clinical diagnosis and measurement of STAS. This process provides a complementary approach for the diagnosis and analysis of STAS by junior pathologists or pathology departments in hospitals with limited medical resources. The process includes: patients undergo pathological slide preparation and scanning at the hospital; subsequently, primary pathologists and DAEM independently perform STAS diagnosis on FS and PS, respectively. For WSIs containing STAS, pathologists can utilize the platform to measure the distance from STAS dissemination foci to the primary tumor. The initial DAEM diagnosis is then reviewed by primary pathologists, with final decisions and detailed pathological diagnostic reports provided by senior pathologists. Finally, thoracic surgeons determine the most appropriate surgical approach (e.g., lobectomy, sublobar resection), while oncologists develop subsequent treatment plans to ensure that patients receive personalized, precise care.